\documentclass[10pt,twocolumn,letterpaper]{article}

\usepackage[pagenumbers]{wacv} 

\usepackage{multirow}
\usepackage{graphicx}
\usepackage[table,xcdraw]{xcolor}

\usepackage[utf8]{inputenc}
\usepackage[T1]{fontenc}
\usepackage{lmodern}

\usepackage[most]{tcolorbox}

\usepackage{verbatim}
\usepackage{stfloats} 
\usepackage{xcolor}            
\usepackage{listings}

\definecolor{myblue}{RGB}{220,245,245}

\newcounter{promptctr}
\renewcommand{\thepromptctr}{Prompt~\Alph{promptctr}}

\tcbset{
  bw:domain/.style={
    colback=white,
    colframe=black!40,
    coltitle=black,
    fonttitle=\bfseries\sffamily,
    colbacktitle=myblue,
    boxrule=0.5pt,
    titlerule=0pt,
    arc=2pt,
    left=4pt,right=4pt,top=4pt,bottom=4pt,
    toptitle=2pt,bottomtitle=2pt,
  }
}

\newcommand{\mytcbinput}[5]{
  \refstepcounter{promptctr}
  \label{#5}
  \begin{tcolorbox}[title={\thepromptctr: #2},#4]
    \lstinputlisting{#1}
  \end{tcolorbox}%
}

\newcommand{\mytcbinputwide}[5]{
  \begin{figure*}[t]
    \centering
    \refstepcounter{promptctr}
    \label{#5}
    \begin{tcolorbox}[title={\thepromptctr: #2},#4,width=\textwidth,enhanced]
      \lstinputlisting{#1}
    \end{tcolorbox}
    \vspace{-4pt}
  \end{figure*}
}

\newcommand{\promptref}[1]{%
  \hyperref[#1]{\ref*{#1}}%
}

\newcommand{\promptrefp}[1]{%
  \hyperref[#1]{\ref*{#1} (p.~\pageref*{#1})}%
}

\newcommand{\method}{\textsc{Perceptual Observatory} }

\definecolor{wacvblue}{rgb}{0.21,0.49,0.74}
\usepackage[pagebackref,breaklinks,colorlinks,allcolors=wacvblue]{hyperref}

\def\wacvPaperID{3314} 
\def\confName{WACV}
\def\confYear{2026}

\title{The Perceptual Observatory \\ Characterizing Robustness and Grounding in MLLMs}

\author{
Tejas Anvekar\thanks{contributed equally} \quad 
Fenil Bardoliya\footnotemark[1] \quad Pavan K. Turaga \quad 
Chitta Baral \quad 
Vivek Gupta \\
Arizona State University \\
\texttt{\{tanvekar, fbardoli, pturaga, chitta, vgupt140\}@asu.edu} \\
\href{https://coral-lab-asu.github.io/PerceptualObservatory/}{https://coral-lab-asu.github.io/PerceptualObservatory/}
}

\begin{document}
\maketitle
\begin{abstract}
Recent advances in multimodal large language models (MLLMs) have yielded increasingly powerful models, yet their perceptual capacities remain poorly characterized. In practice, most model families scale language component while reusing nearly identical vision encoders (e.g., Qwen2.5-VL 3B/7B/72B), which raises pivotal concerns about whether progress reflects genuine visual grounding or reliance on internet-scale textual world knowledge. Existing evaluation methods emphasize end-task accuracy, overlooking robustness, attribution fidelity, and reasoning under controlled perturbations. We present The \method, a framework that characterizes MLLMs across verticals like: (i) simple vision tasks, such as face matching and text-in-vision comprehension capabilities; (ii) local-to-global understanding, encompassing image matching, grid pointing game, and attribute localization, which tests general visual grounding. Each vertical is instantiated with ground-truth datasets of faces and words, systematically perturbed through pixel-based augmentations and diffusion-based stylized illusions. The \method moves beyond leaderboard accuracy to yield insights into how MLLMs preserve perceptual grounding and relational structure under perturbations, providing a principled foundation for analyzing strengths and weaknesses of current and future models.
\end{abstract}

\section{Introduction}
\label{sec:intro}

\begin{figure}[!t]
    \centering
    \includegraphics[width=0.85\linewidth]{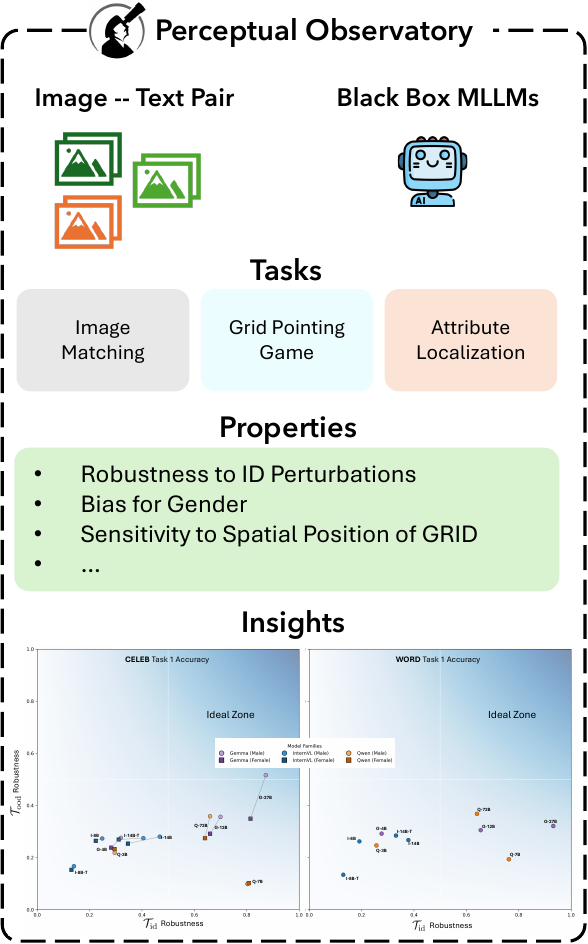}
    \caption{Overview of The \method and how it solicits understanding of opaque MLLMs perceptual understanding by measuring properties motivated by human visual perception and robustness against multiple axes. We illustrate the framework for properties revealing true perceptual understanding of MLLMs} \label{fig:main}
\vspace{-1.5em}
\end{figure}

Multimodal Large Language Models (MLLMs) are ubiquitous for tasks such as captioning, VQA, OCR-centric reasoning, document understanding, accessibility, robotics, and multi-image dialogue~\cite{blip2,flamingo,llava,gpt4v, palm-e, instructblip}. Public leaderboards (e.g., MMBench, MMMU/Pro; TextVQA; VizWiz; SEEDBench; POPE; MATHVista) mostly report end-task accuracy~\cite{mmbench,mmmu-pro,textvqa,vizwiz, seedbench, seedbench2, seedbench2plus, pope, mathvista}. However, outstanding benchmark performance does not guarantee robust \emph{perception} -- defined as the fundamental ability to faithfully understand and interpret visual details, maintain object identity, and spatially ground independent of linguistic reasoning. Without this, models can exploit textual priors, miss identity under perturbations, or fail to localize evidence. 

Modern MLLMs scale the \emph{language} side while leaving vision encoders frozen or lightly adapted via compact bridges (Qwen2.5-VL-family, Gemma3-family, Q-Former, Perceiver resampler, MLP/linear projectors) \cite{qwen2.5-VL,gemma_2025,flamingo,llava}. This raises the question of whether the gains are due to better \emph{visual} or better \emph{textual} capabilities? Decades of vision research warns that models can rely on shortcuts; language priors in VQA or texture bias in CNNs masks poor perceptual grounding~\cite{vqa-vision-matter, dontlook, cnnbias}. Furthermore, while web-scale pretraining increasingly obscures the boundary between In-distribution (ID) and out-of-distribution (OOD) data, foundational robustness studies demonstrate that accuracy can precipitously decline under even modest corruptions or distribution shifts~\cite{imagenet-c,imagenet-a, objectnet}

Based on human-cognitive behaviour, where perception remains robust across stylistic variations and environmental noise~\cite{gestaltpsych1, gestaltpsych2, humans_vs_dnn, tenenbaum_machines_think_people}, we probe the depth of machine \emph{seeing} against these biological standards. We then ask: (Q1) Do MLLMs \emph{preserve identity} under content-preserving ID corruptions and under OOD \emph{stylized images?} (Q2) Are predictions \emph{positional-invariant} when the same content moves in a grid? (Q3) Do models \emph{ground} attributes where they belong, and does giving hints improve transfer? (Q4) Does \emph{scaling} primarily on the language side yield monotonic perceptual gains when the vision encoder is fixed? (Q5) Does enabling \texttt{<think>} mode materially facilitate perception, or just the narrative? (Q6) Are there \emph{fairness} gaps across subpopulations (e.g., gender, race, lighting, texture) under shifts?

For addressing the aforementioned research questions, we introduce The \method, a holistic evaluation suite that measures \emph{how} MLLMs see. We probe with (i) ID augmentations and (ii) OOD \emph{stylized illusions}~\cite{hidden-in-plain-sight} images produced by diffusion with spatial control (Stable/Latent Diffusion {+} ControlNet) that alter appearance while preserving layout, letting us disentangle perception from priors~\cite{ldm,controlnet}. Tasks target complementary skills: identity matching (robustness to perturbations vs. distractors), grid pointing game (spatial invariance), and attribute localization for semi and fully guided settings~\cite{emdm} towards common-sense reasoning~\cite{commonsense} assessment. We summarize our contributions as follows:

\begin{itemize}
  \item We propose The \method: A principled framework that evaluates perceptual robustness and vision-language grounding beyond tradition benchmark performance, highlighting whether failures stem from visual or textual capabilties.
  \item We consolidate simple, interpretable properties of MLLMs like identity robustness, spatial invariance, attribution fidelity, fairness gap, scale consistency, and effects of \texttt{<think>} mode to reveal \emph{how} answers are grounded.
  \item To enable further research in this area, we also provide a scalable pipeline to generate ID corruptions and OOD \emph{stylized illusions} (diffusion{+}ControlNet) that preserve spatial layout while confounding appearance.
  \item Finally, we provide a comprehensive analysis of three leading open-source MLLM families. We demonstrate that scaling the language model without proportional adaptation of the vision encoder results in systematic robustness gaps under distribution shifts, thereby pinpointing the methodological bottlenecks that future research must address.
\end{itemize}

\section{Related Works}
With the recent wave of MLLM families such as Qwen2.5-VL~\cite{qwen2.5-VL}, Gemma3~\cite{gemma_2025}, InternVL3.5~\cite{internvl}, etc., has dramatically pushed the boundaries of visual perception. The large-scale models have frozen or lightly adapted vision backbones such as ViT~\cite{vit}, SigLIP 2~\cite{siglip}, CLIP~\cite{clip}. The early evaluation of these models has emphasized end-task accuracy.  Benchmarks such as MMBench~\cite{mmbench} and MMMU~\cite{mmmu} extend text-centric evaluation to vision language understanding, offering huge collections of diverse QAs (c.f. MMLU~\cite{mmlu}). Yet, these efforts lack perceptual understanding with language priors, leading to the question of whether the high scores on the benchmarks arise from the visual grounding or from the textual reasoning.

The computer vision community has long highlighted the fragility of models under distribution shifts~\cite{imagenet-r, imagenet-c}. Analogous concerns have emerged for MLLMs. Experiments in abstract shape recognition show that VLMs often rely on texture or contextual clues rather than true shape understanding ~\cite{hidden-in-plain-sight}. Similarly, ~\cite{IllusionVQA, IllusoryVQA, grounding-visual-illusion, instinctive-bias, IllusionBench+} construct optical illusions and misleading visual scenarios. These works show that MLLMs are easily misled, as they capture end-task accuracy aided by prompting techniques to improve understanding, yet do not close the gap to human performance without explainability. Another flaw is that the models may have already been trained on certain popular illusions, such as Salvador Dali's painting~\cite{dali-painting}. QAs such as CLEVR~\cite{clevr} and Winoground~\cite{winoground} reveal that models fail to reason on spatial relations and subtle changes ~\cite{aaverma}. 

Beyond QA, VLMs may produce correct answers while attending to irrelevant regions, highlighting poor vision-language disentanglement. Thus, robust multimodal understanding requires attribution localization. Recent MLLMs predict bounding boxes for attributes, enabling explicit evaluation, but localization under distribution shifts for perturbations and illusion remains scarce.

While these prior benchmarks demonstrate critical weaknesses -- language-prior exploitation, fragility to corruption, distribution shifts, and poor grounding -- they traditionally examine \textbf{one dimension at a time}. The \method fills this gap by providing a unified, property-driven assessment of MLLMs across robustness, grounding, and spatial reasoning with controlled low-level augmentations and high-level style-transfer illusions with tasks that explicitly measure identity preservation, spatial invariance, and attribution fidelity. Our \textsc{Observatory} yields a foundation for holistic insights for perceptual strength and weaknesses of MLLMs.
\section{Perceptual Observatory}




The \method is a suite of assessments that characterizes multimodal LLMs across four axes: robustness, in-context adaptation, relational vision, and vision-language alignment as summarized in~\autoref{fig:main}. Unlike accuracy-only benchmarks, it examines \emph{how} models perceive: whether they maintain identity under perturbations, transfer grounding across views, resist distractors, preserve spatial structure, or rely disproportionately on textual priors.

The framework is motivated by principles from perception and cognition, including feature integration~\cite{feature-integration} and structure mapping~\cite{structure-mapping}, which emphasize local-to-global organization and relational reasoning. We instantiate the Observatory in two canonical domains, face recognition and text-in-vision. Then expose models to controlled perturbations comprising (i) pixel-based augmentations (blur, jitter, noise, etc) and (ii) style-transfer based augmentations \emph{``illusions''} generated via Diffusion~\cite{diffusion}+ControlNet~\cite{controlnet}.

The \method then evaluates parameter scales, and decoding modes across model families, yielding comprehensive perceptual profiles capturing robustness behavior, fairness gaps, vision-language alignment, and sensitivity to perturbations. These insights enable principled comparison and selection of reliable MLLM candidates.

\subsection{Problem Statement}

\paragraph{MLLM Characterization.}
We aim to evaluate how a pretrained multimodal LLM \(f\) behaves under controlled visual
perturbations. Each sample in our dataset is a tuple \((x, y, b)\), where \(x\) is an
image, \(y\) is its label (identity or word), and \(b\) contains any available
ground-truth attribute boxes. For a perturbation \(t\) drawn from a transformation set
\(\mathcal{T}\), the model is queried on the modified image \(x' = t(x)\). For a given
property \(P\) (e.g., identity matching, attribute localization), we collect the model's
outputs relevant to that property and measure performance with a task-specific metric
\(M\). This formulation is task-agnostic and accommodates robustness, in-context adaptation, relational vision, and vision-language alignment.

\paragraph{Benchmark Datasets.}
We build two datasets with labeled attributes:  
(i) \textbf{CELEB}~\footnote{\href{https://huggingface.co/datasets/tonyassi/celebrity-1000}{HF Dataset}}, a collection of celebrity faces with identity labels and bounding
boxes for eyes, nose, and mouth; and  
(ii) \textbf{WORD}, a set of synthetically rendered ``text'' images with ground-truth labels and bounding boxes marking the text span.

\paragraph{Perturbations.}
Each dataset has two corresponding sets of perturbed images: 
(i) \emph{Augmentations} (\(\mathcal{T}_{\mathrm{id}}\)), consisting of 15 pixel-level
transformations such as blur, jitter, and noise; and  
(ii) \emph{Illusions} (\(\mathcal{T}_{\mathrm{ood}}\)), 15 stylized transformations, which alter appearance while preserving spatial layout.  For each image \(x\), we sample a transformation \(t\) from either set to obtain \(x' = t(x)\). The complete set of inputs considered in our evaluation is
\[
\mathcal{T} = \{\mathrm{Org}\} \cup \mathcal{T}_{\mathrm{id}} \cup \mathcal{T}_{\mathrm{ood}}
\]
where \(\mathrm{Org}\) denotes the unperturbed original image \(x\).

\subsection{In-Context Formulation}
\label{subsec:icl}

We frame all evaluations as in-context prediction. A model \(f\) is conditioned on a
support example \(S\) an image together with a prompt (and, when relevant, text
annotations) and must answer a query \(Q\). Unless otherwise specified, the support is
the original image \(x^{(\mathrm{Org})}\).

\begin{figure}[!ht]
    \centering
    \includegraphics[width=\linewidth]{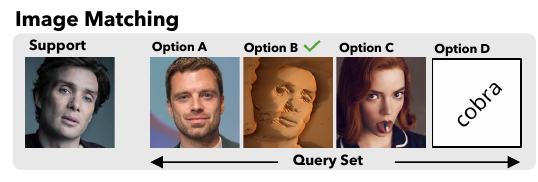}
    \caption{Image Matching: the model selects the candidate, matching the support image. (Supp Sec: \promptref{prompt:image-matching-query} \promptref{prompt:image-matching-support-celeb}, \promptref{prompt:image-matching-support-word}}
    \label{fig:IM}
\end{figure}

\paragraph{Task 1: Image Matching.}
The model is shown a support image and must choose which element in a four-way query set depicts the same entity. As illustrated in~\autoref{fig:IM} (see \textit{Image Matching}), the query set contains four images arranged as \textit{Option A--D} in the figure:

1) the \textbf{correct match} (\emph{option B} in figure), a perturbed version of the support entity (e.g., blurred, stylized, or otherwise transformed); 2) an \textbf{out-of-context} sample drawn from the other domain (face vs.\ text) (\emph{option D}); 3) two \textbf{distractors} (\emph{option A \& C}), chosen as near neighbors CLIP-based nearest faces or words with \(\pm 1\) character edits).

Given support \(S\) and candidates \(\{ \text{A}, \text{B}, \text{C}, \text{D} \}\), the model must output the correct option choice.

\begin{figure}[!ht]
    \centering
    \includegraphics[width=\linewidth]{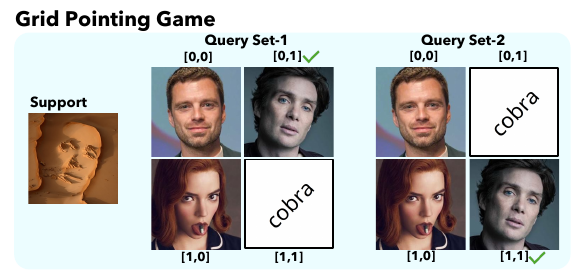}
    \caption{Grid Pointing Game: the model identifies the grid position containing the $\mathrm{Org}$  image. (Supp Sec: \promptref{prompt:gpg-query} \promptref{prompt:gpg-support-celeb}, \promptref{prompt:gpg-support-word}}
    \label{fig:GPG}
\end{figure}

\paragraph{Task 2: Grid Pointing Game.}
The model is given a support image and a \(2 \times 2\) collage (not limitated to) in which the original
image \(x_e^{(\mathrm{Org})}\) (\emph{correct option} \texttt{[0,1]} in query set-1 and \texttt{[1,1]} query set-2) is placed at one of four positions \(\ell\); the remaining three cells contain distractors or out-of-context samples (constructed as in Task~1). As shown in~\autoref{fig:GPG}, the model must point to the location containing the original image by predicting \(\hat{\ell}\). Each entity appears once in every grid position across query sets.



\begin{figure}[!ht]
    \centering
    \includegraphics[width=\linewidth]{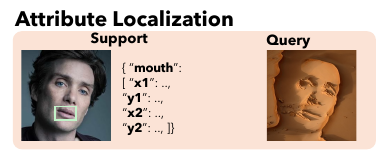}
    \caption{Attribute Localization: the model has to identify attribute information from the support image to the perturbed query. Semi-guided: (Supp Sec: \promptref{prompt:icl-analogy-partial-q3-query-word}, \promptref{prompt:icl-analogy-partial-q3-query-celeb}); Guided: (Supp Sec: \promptref{prompt:icl-analogy-q3-query-word} \promptref{prompt:icl-analogy-q3-support-word}, \promptref{prompt:icl-analogy-q3-query-celeb}, \promptref{prompt:icl-analogy-q3-support-celeb}) }
    \label{fig:AL}
\end{figure}

\paragraph{Task 3: Attribute Localization.}
\label{subsubsec:task3}
For an entity \(e\) with attributes \(\mathcal{A}_e\) and ground-truth boxes
\(\{b_{e,a}\}\), the model is given a support image (with one or more annotated boxes)
and must predict the corresponding attribute boxes \(\hat{b}_{e,a}(t)\) on a perturbed
query image \(x_e^{(t)}\). As shown in~\autoref{fig:AL}, the task evaluates how well the model preserves spatial and structural information under appearance changes.

We consider two variants:
\begin{enumerate}
    \item[a] \textbf{Semi-guided (one-hint):} the support provides a single attribute box,
          and the model must infer the remaining attributes, probing spatial commonsense.
    \item[b] \textbf{Guided (full-hints):} the support provides all attribute boxes, and the model must transfer them to perturbed views probing perceptual consistency.
\end{enumerate}

\subsection{Properties}\label{subsec:properties}

We evaluate both the perceptual robustness of MLLMs and their vision-language alignment. Each property corresponds to an intuitive behavioral goal and a simple quantitative metric.

\textbf{Identity Matching Robustness.} Used for Image Matching and Grid Pointing Game across both datasets. A robust model should preserve entity identity under id and OOD perturbations. We measure the accuracy
drop $\Delta = \mathrm{Acc}(\mathrm{x^{Org}}) - \mathrm{Acc}(x^t)$, where \(t \sim \mathcal{T}_{\mathrm{id}} \cup \mathcal{T}_{\mathrm{ood}}\). Smaller values indicate stronger identity tracking.

\textbf{Gender Bias.} Evaluated on CELEB for all tasks. A fair model should perform similarly on male and female identities. We compute $\mathrm{GAP} = Z_M - Z_F$, using IoU or accuracy depending on the task. Low magnitude of \(\mathrm{GAP}\) indicates gender-neutral behavior.

\textbf{Invariance to Spatial Arrangements.} Specific to the Grid Pointing Game. A position-invariant model should not rely on the grid location of the correct image. For per-position accuracies \(\mathrm{Acc}^\ell\), we
report $\mathrm{Gap}_\ell = \max_\ell \mathrm{Acc}^\ell - \min_\ell \mathrm{Acc}^\ell$. Smaller spreads reflect stronger spatial invariance.

\textbf{Scale Consistency.}
Evaluated across all tasks and datasets. As model size increases within a family, scores \(Z_k\) should improve monotonically with parameter count \(N_k\). We summarize the average gain per parameter doubling. Positive trends indicate scalable perceptual grounding.

\textbf{Thinking Superiority.}
Evaluated across all tasks and datasets. Reasoning-enabled decoding (\texttt{<think>} mode) should enhance perceptual performance. For matched settings, we compute $\Delta^{\text{think}} = Z^{\texttt{<think>}} - Z^{\texttt{base}}$. Positive values indicate that chain-of-thought decoding benefits recognition and grounding.

\textbf{Salient Perceptual Understanding.} Used for Attribute Localization (Task~3). A strong model should preserve salient structure when localizing attributes. \emph{(a) Semi-guided:} we measure the gain from providing one hint, probing spatial commonsense.  \emph{(b) Guided:} we evaluate transfer retention (TR),
\[
\mathrm{TR}(t) =
\frac{\mathrm{mIoU}_{\text{guided}}(t)}
     {\mathrm{mIoU}_{\text{guided}}(\mathrm{Org})},
\]
which tests whether full supervision transfers to perturbed views. High TR indicates stable perceptual layouts under id and OOD shifts.

\section{Experiments}
\label{sec:exp}

\subsection{Dataset}
\label{subsec:dataset}

We construct a two-part benchmark to probe perceptual abilities of multimodal LLMs (MLLMs). \textbf{CELEB} contains 1{,}000 celebrity face images with gold bounding boxes for key features (eyes, nose, mouth), derived from \texttt{MediaPipe}~\cite{mediapipe} and authors manually verified 10\% of the samples and achieved 98\% IoU w.r.t gold. \textbf{WORD} consists of $\sim$267K procedurally rendered words across 21 semantic categories, rendered under diverse fonts, casings, positions, and rotations, yielding $>$1M unique images with exact ground-truth bounding boxes.  

To study robustness, we apply two perturbation families: (1) $\mathcal{T}_\mathrm{id}$ - content-preserving linear augmentations (using Albumentations~\cite{albumentations}), and (2) $\mathcal{T}_\mathrm{ood}$ - style/illusion perturbations using ControlNet~\cite{controlnet} and Stable Diffusion~\cite{diffusion}. Each image has 15 $\mathcal{T}_\mathrm{id}$ variants, 15 $\mathcal{T}_\mathrm{ood}$ variants, and the original, yielding 31K images per dataset and 62K in total.

Further implementation details (augmentation lists, prompt templates, scaling factors) are provided in the supplementary material.

\subsection{Implementation}

\noindent \textbf{MLLMs setup.} We use a variety of MLLMs, including 3 distinct model families: (1) \texttt{Qwen2.5-}\texttt{VL-}\texttt{(3B/7B/72B)-}\texttt{Instruct}~\cite{qwen2.5-VL,qwen2-VL,qwen-VL}, (2) \texttt{Gemma-}\texttt{3-}\texttt{(4B/12B/27B)-}\texttt{Instruct}~\cite{gemma_2025}, and (3) \texttt{InternVL3.5}~\footnote{HF Transformer compatible}\texttt{-(8B/14B)}\texttt{-(Instruct/Thinking)}~\cite{internvl}. The selection was strategically designed to cover a broad spectrum and avoid single evaluation. The key factors included a suite of parameter sizes, distinct model architectures, reasoning capabilities, multi-image inputs, and date of release. 
All experiments were conducted on HPC clusters equipped with NVIDIA $4 \times$H200s with 144GB and $4 \times$H100s with 80GB VRAM, utilizing PyTorch, Huggingface, and the vLLM~\cite{vllm} framework. We maintained a constant \texttt{temperature} of 0.2, \texttt{top\_p} of 0.95, and \texttt{top\_k} of 32 throughout our experimentation.

\begin{table*}[!ht]
\centering
\setlength{\tabcolsep}{4pt}
\caption{Table summarizes robustness of MLLMs for ID vs OOD across both dataset across all task, here Task3(b) is dubbed as Task 3. $\Delta$ refers to difference between current vs smallest among family ex: Qwen7B -- Qwen3B, and also difference between thinking (\texttt{<T>}) vs non-thinking.}
\label{tab:main-table}
\resizebox{\linewidth}{!}{%
\begin{tabular}{
>{\columncolor[HTML]{FFFFFF}}l 
>{\columncolor[HTML]{FFFFFF}}c 
>{\columncolor[HTML]{FFFFFF}}c 
>{\columncolor[HTML]{FFFFFF}}c 
>{\columncolor[HTML]{FFFFFF}}c 
>{\columncolor[HTML]{FFFFFF}}c 
>{\columncolor[HTML]{FFFFFF}}c 
>{\columncolor[HTML]{FFFFFF}}c 
>{\columncolor[HTML]{FFFFFF}}c 
>{\columncolor[HTML]{FFFFFF}}c 
>{\columncolor[HTML]{EFEFEF}}c 
>{\columncolor[HTML]{EFEFEF}}c 
>{\columncolor[HTML]{EFEFEF}}c 
>{\columncolor[HTML]{EFEFEF}}c 
>{\columncolor[HTML]{EFEFEF}}c 
>{\columncolor[HTML]{EFEFEF}}c 
>{\columncolor[HTML]{EFEFEF}}c 
>{\columncolor[HTML]{EFEFEF}}c 
>{\columncolor[HTML]{EFEFEF}}c }
\hline 
 &
  \multicolumn{9}{c}{\cellcolor[HTML]{FFFFFF}\textbf{CELEB}} &
  \multicolumn{9}{c}{\cellcolor[HTML]{EFEFEF}\textbf{WORD}} \\
 &
  \multicolumn{3}{c}{\cellcolor[HTML]{FFFFFF}\textbf{Task1}} &
  \multicolumn{3}{c}{\cellcolor[HTML]{FFFFFF}\textbf{Task2}} &
  \multicolumn{3}{c}{\cellcolor[HTML]{FFFFFF}\textbf{Task3}} &
  \multicolumn{3}{c}{\cellcolor[HTML]{FFFFFF}\textbf{Task1}} &
  \multicolumn{3}{c}{\cellcolor[HTML]{FFFFFF}\textbf{Task2}} &
  \multicolumn{3}{c}{\cellcolor[HTML]{FFFFFF}\textbf{Task3}} \\
\multicolumn{1}{c}{\cellcolor[HTML]{FFFFFF}\textit{\textbf{\#}}} &
  \textbf{$\mathrm{Org}$} &
  \textbf{$\mathcal{T}_{\mathrm{id}}$} &
  \textbf{$\mathcal{T}_{\mathrm{ood}}$} &
  \textbf{$\mathrm{Org}$} &
  \textbf{$\mathcal{T}_{\mathrm{id}}$} &
  \textbf{$\mathcal{T}_{\mathrm{ood}}$} &
  \textbf{$\mathrm{Org}$} &
  \textbf{$\mathcal{T}_{\mathrm{id}}$} &
  \textbf{$\mathcal{T}_{\mathrm{ood}}$} &
  \textbf{$\mathrm{Org}$} &
  \textbf{$\mathcal{T}_{\mathrm{id}}$} &
  \textbf{$\mathcal{T}_{\mathrm{ood}}$} &
  \textbf{$\mathrm{Org}$} &
  \textbf{$\mathcal{T}_{\mathrm{id}}$} &
  \textbf{$\mathcal{T}_{\mathrm{ood}}$} &
  \textbf{$\mathrm{Org}$} &
  \textbf{$\mathcal{T}_{\mathrm{id}}$} &
  \textbf{$\mathcal{T}_{\mathrm{ood}}$} \\ \hline
\multicolumn{19}{c}{\cellcolor[HTML]{E0E2FE}\textit{\textbf{Qwen 2.5 - VL}}} \\ \hline
\textbf{3B} &
  33.66 &
  29.57 &
  {\underline{22.57}} &
  25.00 &
  24.95 &
  24.98 &
  90.57 &
  90.23 &
  \textbf{85.33} &
  21.00 &
  25.66 &
  {\underline{24.66}} &
  25.75 &
  25.53 &
  25.08 &
  97.54 &
  97.48 &
  \textbf{95.35} \\
\textbf{7B} &
  \textbf{78.21} &
  \textbf{80.52} &
  10.00 &
  {\underline{64.75}} &
  {\underline{65.66}} &
  {\underline{29.81}} &
  {\underline{99.65}} &
  {\underline{99.17}} &
  16.52 &
  \textbf{75.00} &
  \textbf{76.26} &
  19.33 &
  {\underline{43.75}} &
  {\underline{47.43}} &
  {\underline{36.33}} &
  \textbf{99.99} &
  \textbf{99.95} &
  {\underline{54.39}} \\
$\Delta$ &
  44.55 &
  50.95 &
  \cellcolor[HTML]{FFD0CC}-12.57 &
  39.75 &
  40.71 &
  \cellcolor[HTML]{FFD0CC}04.83 &
  09.08 &
  08.94 &
  \cellcolor[HTML]{FFD0CC}-68.81 &
  54.00 &
  50.60 &
  \cellcolor[HTML]{FFD0CC}-05.33 &
  18.00 &
  21.90 &
  11.25 &
  \cellcolor[HTML]{FFD0CC}02.45 &
  \cellcolor[HTML]{FFD0CC}02.47 &
  \cellcolor[HTML]{FFD0CC}-40.96 \\
\textbf{72B} &
  {\underline{51.48}} &
  {\underline{65.10}} &
  \textbf{31.61} &
  \textbf{98.25} &
  \textbf{98.66} &
  \textbf{48.40} &
  \textbf{99.99} &
  \textbf{99.93} &
  {\underline{39.65}} &
  {\underline{56.00}} &
  {\underline{64.06}} &
  \textbf{36.80} &
  \textbf{87.75} &
  \textbf{88.40} &
  \textbf{47.06} &
  {\underline{98.64}} &
  {\underline{97.97}} &
  50.55 \\
$\Delta$ &
  17.82 &
  35.53 &
  09.04 &
  73.25 &
  73.71 &
  23.42 &
  09.42 &
  09.70 &
  \cellcolor[HTML]{FFD0CC}-45.68 &
  35.00 &
  38.40 &
  12.14 &
  62.00 &
  62.87 &
  21.98 &
  \cellcolor[HTML]{FFD0CC}01.10 &
  \cellcolor[HTML]{FFD0CC}00.49 &
  \cellcolor[HTML]{FFD0CC}-44.80 \\ \hline
\multicolumn{19}{c}{\cellcolor[HTML]{ECF4FF}\textit{\textbf{Gemma 3}}} \\ \hline
\textbf{4B} &
  24.75 &
  30.09 &
  25.67 &
  45.00 &
  47.10 &
  28.85 &
  \textbf{99.83} &
  \textbf{99.72} &
  \textbf{61.92} &
  23.00 &
  27.66 &
  29.20 &
  46.25 &
  45.56 &
  29.56 &
  99.99 &
  99.99 &
  \textbf{99.69} \\
\textbf{12B} &
  {\underline{55.44}} &
  {\underline{67.98}} &
  {\underline{32.34}} &
  \textbf{84.75} &
  \textbf{84.44} &
  \textbf{40.03} &
  99.54 &
  {\underline{98.47}} &
  {\underline{51.86}} &
  {\underline{66.00}} &
  {\underline{65.46}} &
  {\underline{30.53}} &
  {\underline{64.50}} &
  {\underline{64.93}} &
  \textbf{40.68} &
  {\underline{99.99}} &
  {\underline{99.99}} &
  {\underline{99.28}} \\
$\Delta$ &
  30.69 &
  37.89 &
  \cellcolor[HTML]{FFD0CC}06.67 &
  39.75 &
  37.34 &
  11.18 &
  \cellcolor[HTML]{FFD0CC}-00.29 &
  \cellcolor[HTML]{FFD0CC}-01.25 &
  \cellcolor[HTML]{FFD0CC}-10.06 &
  43.00 &
  37.80 &
  \cellcolor[HTML]{FFD0CC}01.33 &
  18.25 &
  19.37 &
  11.12 &
  00.00 &
  00.00 &
  \cellcolor[HTML]{FFD0CC}-00.41 \\
\textbf{27B} &
  \textbf{78.21} &
  \textbf{84.35} &
  \textbf{43.16} &
  {\underline{71.50}} &
  {\underline{68.63}} &
  {\underline{31.40}} &
  {\underline{99.58}} &
  97.38 &
  25.86 &
  \textbf{96.00} &
  \textbf{93.20} &
  \textbf{32.13} &
  \textbf{69.00} &
  \textbf{67.50} &
  {\underline{38.46}} &
  \textbf{99.99} &
  \textbf{99.99} &
  80.27 \\
$\Delta$ &
  53.46 &
  54.26 &
  17.49 &
  26.50 &
  21.53 &
  \cellcolor[HTML]{FFD0CC}02.55 &
  \cellcolor[HTML]{FFD0CC}-00.25 &
  \cellcolor[HTML]{FFD0CC}-02.34 &
  \cellcolor[HTML]{FFD0CC}-36.06 &
  73.00 &
  65.54 &
  \cellcolor[HTML]{FFD0CC}02.93 &
  22.75 &
  21.94 &
  \cellcolor[HTML]{FFD0CC}08.90 &
  00.00 &
  00.00 &
  \cellcolor[HTML]{FFD0CC}-19.42 \\ \hline
\multicolumn{19}{c}{\cellcolor[HTML]{FEF5DC}\textit{\textbf{InternVL 3.5}}} \\ \hline
\textbf{8B} &
  {\underline{18.81}} &
  23.56 &
  {\underline{26.86}} &
  81.75 &
  79.71 &
  36.56 &
  \textbf{99.99} &
  \textbf{99.99} &
  \textbf{66.84} &
  13.00 &
  19.11 &
  26.20 &
  {\underline{47.00}} &
  44.53 &
  27.31 &
  \textbf{99.99} &
  \textbf{99.99} &
  \textbf{99.64} \\
-- \texttt{<T>} &
  03.96 &
  13.53 &
  15.97 &
  29.25 &
  31.73 &
  26.66 &
  {\underline{96.99}} &
  {\underline{95.50}} &
  17.79 &
  08.00 &
  12.99 &
  13.40 &
  06.25 &
  06.33 &
  09.56 &
  72.99 &
  66.76 &
  47.82 \\
$\Delta$ &
  \cellcolor[HTML]{FFD0CC}-14.85 &
  \cellcolor[HTML]{FFD0CC}-10.03 &
  \cellcolor[HTML]{FFD0CC}-10.89 &
  \cellcolor[HTML]{FFD0CC}-52.50 &
  \cellcolor[HTML]{FFD0CC}-47.98 &
  \cellcolor[HTML]{FFD0CC}-09.90 &
  \cellcolor[HTML]{FFD0CC}-03.00 &
  \cellcolor[HTML]{FFD0CC}-04.49 &
  \cellcolor[HTML]{FFD0CC}-49.05 &
  \cellcolor[HTML]{FFD0CC}-05.00 &
  \cellcolor[HTML]{FFD0CC}-06.12 &
  \cellcolor[HTML]{FFD0CC}-12.80 &
  \cellcolor[HTML]{FFD0CC}-40.75 &
  \cellcolor[HTML]{FFD0CC}-38.20 &
  \cellcolor[HTML]{FFD0CC}-17.75 &
  \cellcolor[HTML]{FFD0CC}-27.00 &
  \cellcolor[HTML]{FFD0CC}-33.23 &
  \cellcolor[HTML]{FFD0CC}-51.82 \\
\textbf{14B} &
  27.72 &
  \textbf{40.66} &
  26.66 &
  \textbf{98.50} &
  \textbf{97.91} &
  {\underline{47.08}} &
  49.41 &
  53.39 &
  {\underline{52.34}} &
  \textbf{34.00} &
  \textbf{37.86} &
  {\underline{26.73}} &
  \textbf{71.25} &
  \textbf{72.40} &
  {\underline{43.55}} &
  98.61 &
  {\underline{98.12}} &
  {\underline{95.42}} \\
-- \texttt{<T>} &
  \textbf{27.72} &
  {\underline{35.77}} &
  \textbf{27.19} &
  {\underline{89.25}} &
  {\underline{92.53}} &
  \textbf{51.18} &
  85.04 &
  81.49 &
  11.17 &
  {\underline{27.00}} &
  {\underline{33.13}} &
  \textbf{28.46} &
  44.00 &
  {\underline{44.59}} &
  \textbf{49.41} &
  {\underline{99.22}} &
  97.30 &
  92.89 \\
$\Delta$ &
  \cellcolor[HTML]{FFD0CC}00.00 &
  \cellcolor[HTML]{FFD0CC}-04.89 &
  \cellcolor[HTML]{FFD0CC}00.53 &
  \cellcolor[HTML]{FFD0CC}-09.25 &
  \cellcolor[HTML]{FFD0CC}-05.38 &
  \cellcolor[HTML]{FFD0CC}04.10 &
  35.63 &
  28.10 &
  \cellcolor[HTML]{FFD0CC}-41.17 &
  \cellcolor[HTML]{FFD0CC}-07.00 &
  \cellcolor[HTML]{FFD0CC}-04.73 &
  \cellcolor[HTML]{FFD0CC}01.73 &
  \cellcolor[HTML]{FFD0CC}-27.25 &
  \cellcolor[HTML]{FFD0CC}-27.81 &
  \cellcolor[HTML]{FFD0CC}05.86 &
  \cellcolor[HTML]{FFD0CC}00.61 &
  \cellcolor[HTML]{FFD0CC}-00.82 &
  \cellcolor[HTML]{FFD0CC}-02.53 \\ \hline
\multicolumn{19}{c}{\cellcolor[HTML]{DDF8EB}\textit{\textbf{Human}}} \\ \hline
 &
  100 &
  100 &
  89.11 &
  100 &
  100 &
  87.55 &
  95.66 &
  93.22 &
  81.88 &
  100 &
  100 &
  83.86 &
  100 &
  100 &
  79.98 &
  99.99 &
  99.99 &
  99.99 \\ \hline 
\end{tabular}%
}
\end{table*}
\section{Results \& Discussion}
\label{sec:results}

As a preview of our results that we will describe in detail, we establish four consistent themes across properties defined in \S\ref{subsec:properties}: (1) \textbf{WORD} tasks are near-saturated in-distribution and retain accuracy under shift, whereas \textbf{CELEB} tasks degrade sharply out-of-distribution. (2) Scaling primarily benefits OCR, pointing, and guided localization, but does not guarantee robustness to identity-preserving perturbations. (3) \textbf{LM-side capacity} (decoder depth/width, projector dimension) drives most of the gains, since the vision encoder is held fixed. (4) Decode-time reasoning (\texttt{<think>}) enhances clean/ID performance but reduces transfer retention on faces. Humans achieve near-ceiling accuracies, highlighting that gaps are model-driven rather than dataset artifacts.

\subsection{Identity Matching Robustness}
\label{subsec:robustness}
\begin{figure}[ht]
  \centering
  \includegraphics[width=\linewidth]{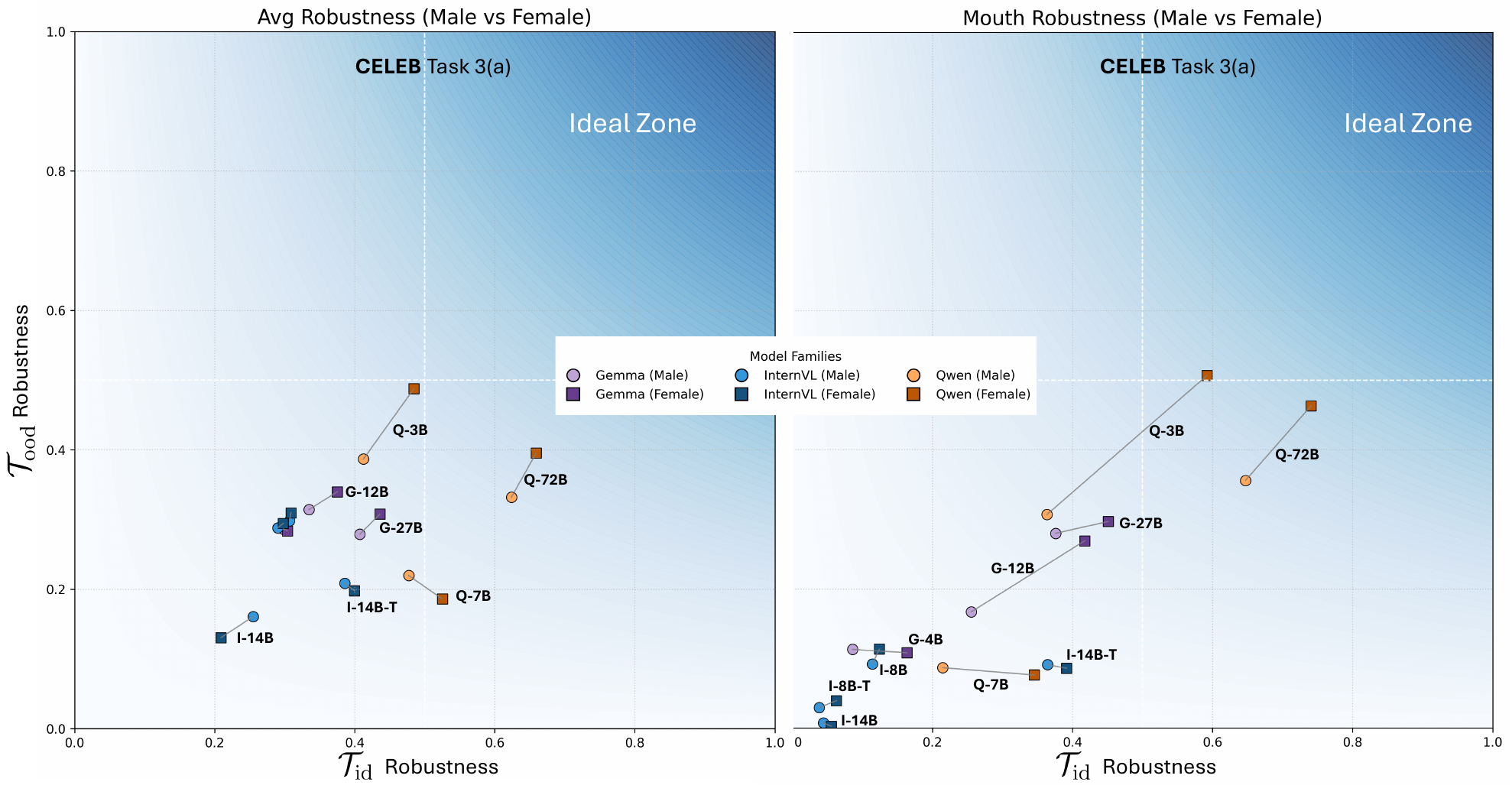}
  \caption{Left figure demonstrates Multidimensional Insights for Task 3(a) across all attributes (eyes, nose, mouth), gender gap, robustness on ID vs.\ OOD for CELEB . Whereas right figure provides fine-grain insights for a specific attribute: \texttt{mouth}.}
  \label{fig:task3a}
\end{figure}

~\autoref{tab:main-table} depicts $\Delta$ under $\mathcal{T}_\mathrm{id}$ and $\mathcal{T}_\mathrm{ood}$. Three robust trends emerge:  
(i) ID augmentations (blur, noise, etc.) produce negligible loss and occasionally improve accuracy.  
(ii) OOD illusions disproportionately harm \emph{mid-scale} models (7--14B), along with larger MLLMs (Qwen-72B, Gemma-27B) fail to retain higher robustness. (iii) Robustness is non-monotonic: e.g., Gemma-12B is more brittle than Gemma-4B, highlighting that methodological flaws can outweigh scale.  Grounding in ~\autoref{tab:main-table} similar trends can be inferred for WORD across Org/ID/OOD. Human annotators exceed $95\%$ across all conditions, establishing an empirical ceiling.

\begin{figure*}[ht]
  \centering
  \includegraphics[width=\linewidth]{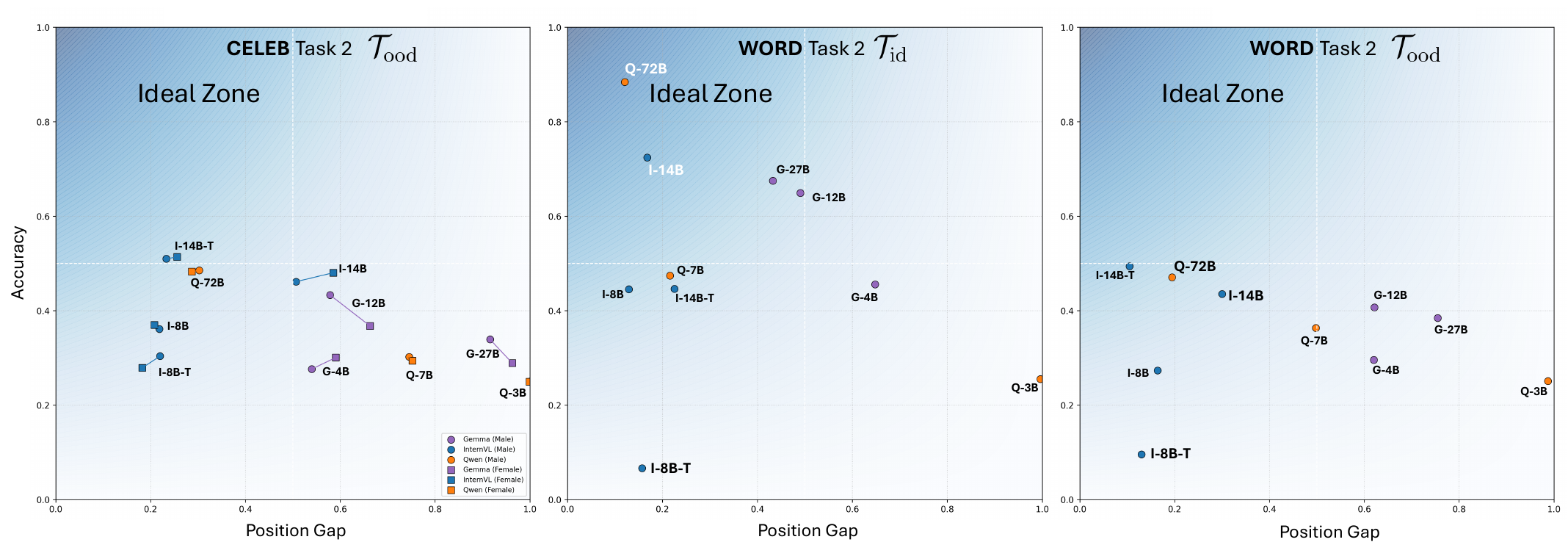}
  \caption{Multidimensional insights for Task 2, accuracy vs.\ position gap across perturbation, across datasets, this figure reveals majority of the models suffer under OOD setting with high gender gap suggesting sensitivity to grid position.}
  \label{fig:task2}
\end{figure*}

\subsection{Invariance to Spatial Arrangements}
\label{subsec:spatial}

In the Grid Pointing Game, \autoref{fig:task2} reveals insights that several models show pronounced positional biases, with gap spreads (\S~\ref{subsec:properties})  exceeding $approx.$ $50-90\%$ for small--to--large MLLMs. For CELEB, only \texttt{InternVL-3.5-8B-}\texttt{thinking} has the least position bias, but also has the worst accuracy, depicting that, \emph{thinking} doesn't facilitate in \emph{seeing}. Even for simple WORD, models (\texttt{Qwen2.5-VL-72B} \& \texttt{InternVL-3.5-14B}) which seem to be in ``ideal zone'' tend to fail as we switch from ID to OOD this suggest stylistic change were not incorporated by language understanding, as vision encoder was never aligned jointly. Larger decoders reduce $\mathrm{Gap}_\ell$ on WORD but only partially on CELEB, confirming that  the encoder, regulate spatial invariance.

\subsection{Gender Bias in CELEB}
\label{subsec:bias}

\autoref{fig:task3a} (a) and \autoref{fig:task2} provide a visualization to understand gender bias (depicted as line between round and square marker; \emph{larger line depicts huge gender bias}) in MLLMs along with assessing other axes like robustness for OOD vs.\ ID. As depicted in \autoref{fig:task2}, models like \texttt{InternVL} and \texttt{Gemma} have gender bias, especially for \texttt{Gemma-12B} where-in the model goes from low position gap to worst when gender is changed from male to female.

One can observe in ~\autoref{fig:task3a} that for Task~3(a): Semi-guided attribution task, most models have gender bias, when analyzing  with fine-grain lens, example: just for Attribute: \texttt{mouth}; models become much worse (Ex: \texttt{Qwen2.5-VL-3B} \& \texttt{72B}, \texttt{Gemma-12B}). Due to the visual backbone being unchanged across sizes, we hypothesize that, gains arise from stronger cross-modal calibration in the textual space rather than visual. Nonetheless, asymmetries persist without explicit debiasing.

\subsection{Scale Consistency}
\label{subsec:scaling}
\autoref{tab:main-table} summarizes scale-consistency, i.e. just scaling the language model may not be the right way to improve performance, as \texttt{Qwen}(\texttt{3B}$\rightarrow$\texttt{7B}), \texttt{InternVL}(\texttt{8B}$\rightarrow$\texttt{14B}) , and \texttt{Gemma}(\texttt{4B}$\rightarrow$\texttt{12B}$\rightarrow$\texttt{27B}) performance is improved drastically on Task~1: ($\mathrm{Org}, \mathcal{T}_\mathrm{id}$) for CELEB \& WORD, whereas further scaling collapses the \texttt{Qwen} to \texttt{72B} performance. Contrary, for Task~1: ($\mathcal{T}_\mathrm{ood}$) the model performance is either half of the $\mathrm{Org}$, $\mathcal{T}_\mathrm{id}$; or its smaller variant. This suggests that the models rely on the training \emph{``world knowledge''} rather than focusing on \emph{``visual cues''}. This clearly necessitates the need for joint alignment of both vision-encoder and language-decoder for scaling.

\subsection{Task-Level Grounding}
\textbf{Task-3 (Attribute Localization).} \autoref{fig:task3a} Task~3(a) shows no models are even close to the Ideal Zone (High IoU for $\mathcal{T}_{\mathrm{id}}$, $\mathcal{T}_{\mathrm{ood}}$). Even from \autoref{tab:main-table} Task~3(b), a simply cognitive task of attribute transcription, larger models perform poorly compared to smaller counter parts.  \textbf{Mouth Localization}. \autoref{fig:task3a} highlights that ID distributions peak at high IoU, but OOD shifts the distribution with a low-IoU for almost all models, but especially all \texttt{InternVL} variants suffer from OOD distribution change. This corroborates our spatial-invariance findings.  

\subsection{Thinking Superiority}
\label{subsec:thinking}


\begin{figure}[ht]
  \centering
  \includegraphics[width=\linewidth]{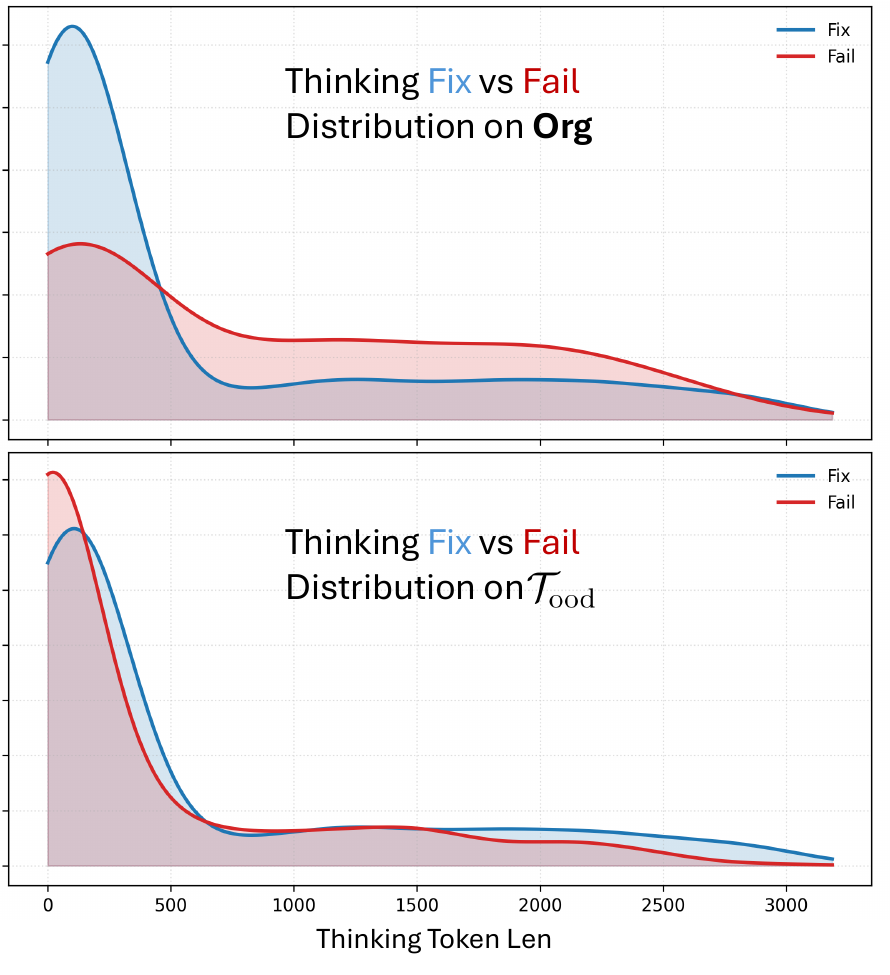}
  \caption{Celeb chain length vs.\ outcome. Histogram (log-$y$) of \texttt{<think>} token length for cases where reasoning \emph{fixes} vs.\ \emph{fails}. Top: $\mathrm{Org}$ fixes vs.\ fails. Bottom: $\mathcal{T}_{\mathrm{ood}}$  fixes vs.\ fails.}
  \label{fig:thinking_length}
\end{figure}

We assess, whether \texttt{<think>} mode actually thinks? \autoref{tab:main-table} Task~1,2,3(b), \texttt{InternVL-3.5-8B}\texttt{-thinking} fails on all tasks compared to its non-thinking variant, as highlighted in red color. \texttt{InternVL-3.5-14B}\texttt{-thinking} also follows similar trends of poor performance compared to the non-thinking variant. \autoref{fig:task3a} shows that for attribute: mouth, the thinking variant of \texttt{InternVL} has the lowest robustness compared to other non-thinking models. Moreover, \autoref{fig:task2} \texttt{InternVL-3.5-8B}\texttt{-thinking} has the lowest position-gap for WORD, but of no use as the accuracy is very poor (below $10\%$). Also, for CELEB, it has more gender-bias compared to its non-thinking counterpart.

\paragraph{Reasoning chain length.} \autoref{fig:thinking_length} shows that on $\mathrm{Org}$, successful fixes concentrate at short reasoning chain lengths, while failures still do occur when operating with chain length $\sim2000$ tokens. Contrary, for $\mathcal{T}_{\mathrm{ood}}$, rare fixes appear in longer-chain tails, and the model most of the time gives up during the early stage of reasoning with high confidence, suggesting over-reliance on textual knowledge compared to visual.

\subsection{Salient Perceptual Understanding}
\label{subsec:saliency}
From ~\autoref{fig:task3a}, we observe that on Task~3(a), almost all models suffer in spatial-common sense understanding, i.e, given ``nose'' or ``top-left-corner'' coords, models struggle to identify other attributes which are spatially very near.  
Transfer retention drops for simple cognitive tasks like guided transcription and semi-guided attribute localization on CELEB \& WORD for $\mathcal{T}_{\mathrm{ood}}$. We demonstrate other multi-dimensional vulnerabilities like gender-bias, spatial-invariance, robustness to OOD samples, scaling effects, and true performance of \texttt{<think>} mode of current MLLMs using \autoref{fig:task2} \& \autoref{fig:task3a}.

\subsection{Human Baseline}
\label{subsec:human}
To contextualize model performance, we conducted a human study on both \textbf{CELEB} and \textbf{WORD}.  
For each dataset, 100 samples were randomly chosen and evaluated across all three tasks (\S\ref{subsec:icl}) by two annotators, achieving an average inter-annotator agreement of 94.5\%.  

As shown in ~\autoref{tab:main-table}, humans achieved near-perfect accuracies ($>95\%$) on identity and spatial tasks, with only mild degradation under $\mathcal{T}_{\mathrm{ood}}$ perturbations.  
On attribute localization, annotators retained high performance ($81\%$ mIoU in the most challenging guided-perturbation setting), even in the semi-guided case.  

These results establish the empirical upper bound: the tasks are perceptually tractable for humans, and gaps in robustness, spatial invariance, or grounding can be attributed to limitations of current MLLMs rather than dataset artifacts.

\section{Limitations}
\label{sec:limitations}
While the \method provides a principled framework for assessing MLLMs, several limitations remain. 
First, the evaluation is restricted to two domains (faces and synthetic words), limiting conclusions 
about broader perceptual generalization. Second, human annotations for illusions were verified only 
on a subset, and baselines were derived from a small sample with few annotators, which constrains 
statistical robustness. Third, fairness analysis focused on gender, leaving other social factors such 
as skin tone unexplored. Fourth, experiments were limited to open-source models for transparency and 
feasibility, excluding closed-source systems. These choices were deliberate to ensure tractability 
and interpretability, but expanding datasets, annotations, social dimensions, and model coverage 
remains an important direction for future work.


\section{Conclusion \& Future Work}

This work introduced The \method, a principled framework for holistic evaluation of visual capabilities of MLLMs, by combining controlled pixel-based augmentations along with diffusion-based styled illusions, and by evaluating tasks spanning identity matching, grid-based spatial reasoning, and attribute localization. This Observatory moves beyond traditional leaderboard benchmarks. Our proposed property and insights lay the foundation for robustness, failures arising from vision encoders, language decoder scaling, and reasoning capabilities that reveals inherent flaws in grounding and fairness across model families and sizes. We observed, scaling language decoders does not guarantee monotonic gains in visual grounding and hinders the visual understanding under OOD distribution shifts. These insights showcase the importance of evaluating how models “see”, not only how well they answer, and provide actionable insights for designing next-generation multimodal models.

To extend the impact of the \method, we aim to broaden the dataset scope beyond celebrity faces and synthetic words to more diverse visual domains. This will enable more comprehensive and holistic evaluation of multimodal models' visual strengths and weaknesses. Furthermore, the expanded dataset will serve as a foundation for joint vision-language alignment. Instead of scaling only the language component, we propose a joint optimization framework that simultaneously scales both vision and language components. Leveraging reinforcement learning for post-training, we will use the property-based metrics defined in this work as rewards. This approach ensures that vision is given equal importance, potentially improving alignment and robustness.

Additionally, we identify the need for a deeper evaluation of reasoning chains in MLLMs. While our analysis touched on reasoning-enabled decoding, there is currently no standard metric for evaluating reasoning quality. Future work will develop and incorporate such metrics to provide a clearer understanding of how reasoning chains contribute to model performance and robustness.

\label{sec:conclusion}

\section{Acknowledgement}
We thank the Complex Data Analysis and Reasoning Lab at Arizona State University for computational support. The work was partially supported by NSF grant 2323086.

{
    \small
    \bibliographystyle{ieeenat_fullname}
    \bibliography{main}
}

\maketitlesupplementary

\section{Dataset Details}
\label{supsec:dataset_details}

\subsection{CELEB}
\label{supsec:celeb}
We sample 1{,}000 celebrity face images for facial feature attribution. Bounding boxes for left/right eyes, nose, and mouth are computed using \texttt{MediaPipe}. To validate reliability, the first and second authors manually annotated 10\% of images, achieving 98\% IoU with \texttt{MediaPipe} outputs. Hence, we treat \texttt{MediaPipe}-derived boxes as gold annotations.

\subsection{WORD}
\label{supsec:word}
We collect $\sim$267K unique words across 21 semantic categories (\emph{Computer Science, Cities, People, Food, Politics, Abuse}, etc.). Word length $l \in [2,10]$ with $\mathbb{E}[l]\!\approx\!4.8$. Each word is rendered under:
\[
\mathcal{F}\times\mathcal{C}\times\mathcal{P}\times\mathcal{R},
\]
with $\mathcal{F}=\{CourierNew,..., TimesNewRoman\}$ (fonts),  
$\mathcal{C}=\{\text{upper}, \text{lower}, \text{camel}\}$ (casings),  
$\mathcal{P}=\{\text{center}, \text{top}, \text{bottom}\}$ (positions),  
$\mathcal{R}=\{-45^\circ,0^\circ,45^\circ\}$ (rotations).  
Uniform sampling across these factors produces $>$1M rendered images overall. Because WORD is procedurally generated, bounding boxes are exactly known.

\subsection{Perturbations}
\label{supsec:perturb}
We apply two perturbation families:

\paragraph{Linear augmentations ($\mathcal{P}_1$).}  
Implemented with Albumentations~\cite{albumentations}. Each image is augmented by sampling from the set
\[
\resizebox{1\columnwidth}{!}{$
\mathcal{M} =
\left\{
\begin{array}{l}
\texttt{GaussianBlur}(11,11),\; \texttt{MedianFilter}(21), \\
\texttt{ZoomBlur}([1.05,1.07]),\; \texttt{ChromaticAberration}(\pm0.2), \\
\texttt{ISONoise}([0.01,0.05],[0.1,0.5]),\; \texttt{RGBShift}(\pm20), \\
\texttt{Salt\&PepperNoise}([10^{-4},10^{-3}]),\; \texttt{GammaLimit}([80,140]), \\
\texttt{JPEGCompression}([20,50]),\; \texttt{MultiplicativeNoise}([0.9,1.1]), \\
\texttt{Sharpen}(\alpha \in [0.3,0.5]),\; \texttt{GlassBlur}(\sigma=0.3,\Delta=2), \\
\texttt{Posterize}(4\;\text{bits}),\; \texttt{MotionBlur}(7,7), \\
\texttt{GaussianNoise}(\mu=0,\;\sigma \in [0.05,0.1])
\end{array}
\right\}
$}
\]
Thus, $\mathcal{P}_1(x)\sim\mathcal{U}(\mathcal{M})$.

\paragraph{Illusion perturbations ($\mathcal{P}_2$).}  
Following IllusionBench~\cite{hidden-in-plain-sight}, each source image $x_i$ is embedded into a stylized scene using ControlNet~\cite{controlnet} with Stable Diffusion~\cite{diffusion}. Prompts are composed from:
\[
[\texttt{SubjectScene}] \times [\texttt{Style}] \times [\texttt{Light/ColorHighlight}],
\]
where representative values are listed below:

\begin{center}
\resizebox{\linewidth}{!}{
\begin{tabular}{lll}
\toprule
\textbf{Subject Scene} & \textbf{Style} & \textbf{Light/Color Highlight} \\
\midrule
Museum & Cinematic & Dust Motes \\
Rainy Alleyway & Gothic & Neon Glow \\
Forest & Fantasy Art & Golden Hour \\
Desert Dune & Vintage Photo & Pastel Hues \\
Medieval Village & Minimalist & Stark Shadows \\
Ocean & Surrealism & Electric Blue \\
Sunset Beach & Bioluminescent & Crystal Refraction \\
Cozy Cottage & Origami & Venetian Blinds \\
Mountain Range & Dystopian & Hearth Fire \\
Overgrown Ruins & Abstract & Volumetric Rays \\
Starry Night & Painting & Smudged Grays \\
Cloudy & Pixel Art & Pink Cyan \\
\bottomrule
\end{tabular}
}
\end{center}

We apply a negative prompt (\texttt{glitch, low quality}) to suppress artifacts. Control strengths are dataset-dependent:  
WORD: $cn\_scale=1.2,\ guide\_scale=10.5$,  
CELEB: $cn\_scale=3.0,\ guide\_scale=7.5$.

Each final entry is stored as $(x_{ij}, s_j)$, where $s_j$ encodes the sampled scene, style, and lighting.

\subsection*{Final Dataset Size}
\label{supsec:size}
For each dataset (CELEB, WORD), we sample 1{,}000 original images and generate 15 variants with $\mathcal{P}_1$, 15 with $\mathcal{P}_2$, plus the original. This yields:
\[
1000 \times (1 + 15 + 15) = 31{,}000 \;\; \text{images per dataset}.
\]
In total, the benchmark contains \textbf{62{,}000} images.

\section{Prompt Templates}
\label{supsec:prompts}

\mytcbinput{sec/prompts/image-matching-query.tex}{Image Matching Query}{0}{bw:domain}{prompt:image-matching-query}
\mytcbinput{sec/prompts/image-matching-support-celeb.tex}{Image Matching Support (Celeb)}{0}{bw:domain}{prompt:image-matching-support-celeb}
\mytcbinput{sec/prompts/image-matching-support-word.tex}{Image Matching Support (Word)}{0}{bw:domain}{prompt:image-matching-support-word}

\mytcbinput{sec/prompts/gpg-query.tex}{GPG Query}{0}{bw:domain}{prompt:gpg-query}
\mytcbinput{sec/prompts/gpg-support-celeb.tex}{GPG Support (Celeb)}{0}{bw:domain}{prompt:gpg-support-celeb}
\mytcbinput{sec/prompts/gpg-support-word.tex}{GPG Support (Word)}{0}{bw:domain}{prompt:gpg-support-word}

\mytcbinput{sec/prompts/icl-analogy-q3-support-celeb.tex}{Attribution Support (Celeb)}{0}{bw:domain}{prompt:icl-analogy-q3-support-celeb}
\mytcbinput{sec/prompts/icl-analogy-q3-support-word.tex}{Attribution Support (Word)}{0}{bw:domain}{prompt:icl-analogy-q3-support-word}

\mytcbinput{sec/prompts/icl-analogy-q3-query-word.tex}{Attribution Guided Query (Word)}{0}{bw:domain}{prompt:icl-analogy-q3-query-word}

\mytcbinput{sec/prompts/icl-analogy-partial-q3-query-word.tex}{Attribution Semi Guided Query (Word)}{0}{bw:domain}{prompt:icl-analogy-partial-q3-query-word}

\mytcbinputwide{sec/prompts/icl-analogy-q3-query-celeb.tex}{Attribution Guided Query (Celeb)}{0}{bw:domain}{prompt:icl-analogy-q3-query-celeb}
\mytcbinputwide{sec/prompts/icl-analogy-partial-q3-query-celeb.tex}{Attribution Semi Guided Query (Celeb)}{0}{bw:domain}{prompt:icl-analogy-partial-q3-query-celeb}

\end{document}